# Robust Activity Recognition for Adaptive Worker-Robot Interaction using Transfer Learning

Farid Shahnavaz[1], Riley Tavassoli[2], and Reza Akhavian, Ph.D., M. ASCE[3]

[1]Graduate Student, Computational Science Research Group, San Diego State University, San Diego, CA. Email: fshahnavaz2347@sdsu.edu
[2]Visiting Scholar, Dept. of Civil, Construction, and Environmental Engineering, San Diego State University, San Diego, CA. Email: rtavassoli@sdsu.edu
[3]Associate Professor, Dept. of Civil Engineering, San Diego State University, San Diego, CA (corresponding author). ORCID: https://orcid.org/0000-0001-9691-8016. Email: rakhavian@sdsu.edu

## ABSTRACT

Human activity recognition (HAR) using machine learning has shown tremendous promise in detecting construction workers' activities. HAR has many applications in human-robot interaction research to enable robots' understanding of human counterparts' activities. However, many existing HAR approaches lack robustness, generalizability, and adaptability. This paper proposes a transfer learning methodology for activity recognition of construction workers that requires orders of magnitude less data and compute time for comparable or better classification accuracy. The developed algorithm transfers features from a model pre-trained by the original authors and fine-tunes them for the downstream task of activity recognition in construction. The model was pre-trained on Kinetics-400, a large-scale video-based human activity recognition dataset with 400 distinct classes. The model was fine-tuned and tested using videos captured from manual material handling (MMH) activities found on YouTube. Results indicate that the fine-tuned model can recognize distinct MMH tasks in a robust and adaptive manner which is crucial for the widespread deployment of collaborative robots in construction.

## INTRODUCTION

Human activity recognition (HAR) has gained significant traction in recent years due to its potential applications in various fields, including healthcare, sports, security, and construction. In the construction industry, the accurate and real-time recognition of workers' activities is imperative for ensuring safety, improving productivity, and optimizing resource allocation. HAR can be achieved using wearable sensors or vision-based methods, with both approaches showing promising results in human-robot interaction (HRI) research (Liu et al., 2022; Zhang et al., 2017).

Researchers have employed multiple tools, including wearable sensors, cameras, and other types of sensing devices, to detect human activities for HAR in various domains such as healthcare, sports, and construction. Vision-based methods, which involve analyzing visual data from cameras to detect and identify human activities, are one of the most popular approaches for HAR. In the construction domain, Luo et al. developed a vision-based system that uses a single camera to monitor workers' activities and detect unsafe behavior (Luo et al., 2018). Similarly, Escorcia et al.

proposed a system that uses a combination of RGB and depth sensors to recognize construction workers' activities. Another common approach for HAR is using wearable sensors, such as accelerometers and gyroscopes, to capture human movements (Escorcia et al., 2012). Such sensors have been used extensively in construction to monitor workers' activities and detect unsafe behavior. For instance, Kim and Cho. developed a wearable sensor-based system that uses machine learning to recognize construction workers' activities and detect unsafe behavior (K. Kim & Cho, 2020).

However, existing approaches to HAR often fail to capture the heterogeneity of activity types, environments, and subjects. This is because the models are often trained on limited datasets and may not be able to perform well in different environments or with different subjects. This has resulted in machine learning models that lack robustness, adaptability, generalizability, and reconfigurability when applied in conditions different from those in which they were trained (Zhang et al., 2022). This limitation poses a significant challenge for the deployment of collaborative robots in construction, as they require robust and adaptive HAR models.

Transfer learning (TL) is a promising approach that can address this challenge, where knowledge gained from a source domain can be transferred to a target domain to improve performance. Within the construction research domain, precious studies have successfully used TL to detect objects such as guardrails, hard hats, and equipment (H. Kim et al., 2018; Kolar et al., 2018; Shen et al., 2021). Nevertheless, the use of TL for HAR and specifically toward deploying it in worker-robot interaction applications has never been investigated before. This paper presents a TL methodology for activity recognition of construction workers interacting with collaborative robots. To achieve TL in video-based activity recognition, we used X-CLIP (Expanding Contrastive Language-Image Pre-training) (Ni et al., 2022), a model developed by Microsoft that extends the functionality of the original CLIP model by OpenAI (Radford et al., 2021). X-CLIP was specifically designed for video recognition tasks and has demonstrated excellent performance in various video and text-based tasks. By leveraging its powerful multimodal learning capabilities, X-CLIP is expected to provide superior performance in the activity recognition task for construction workers interacting with collaborative robots. In this paper, first, we utilize a pre-trained model from a large-scale video-based HAR dataset, Kinetics-400, which has not been used before in the context of construction activity recognition. Second, we fine-tune the pre-trained model on a small number of construction-specific activities, which require minimal annotation efforts and computational resources, making it more feasible for real-world deployment. Third, we demonstrate the effectiveness of our approach in recognizing manual material handling activities in construction, which is crucial for enabling the deployment of collaborative robots in this domain.

**METHODOLOGY**

The proposed methodology for fine-tuning a general activity recognition model for MMH activities using X-CLIP is shown in Figure 1 and described below.

**Data Collection**. The first step in developing the model is to collect video data of construction MMH activities involving workers. These videos will be used for training and testing the model. The videos should cover a wide range of scenarios, such as different workers, different types of material, and different environments. This resulted in having a diverse dataset of video data covering various scenarios, which is crucial for training and testing the model to develop a construction MMH activity recognition system.

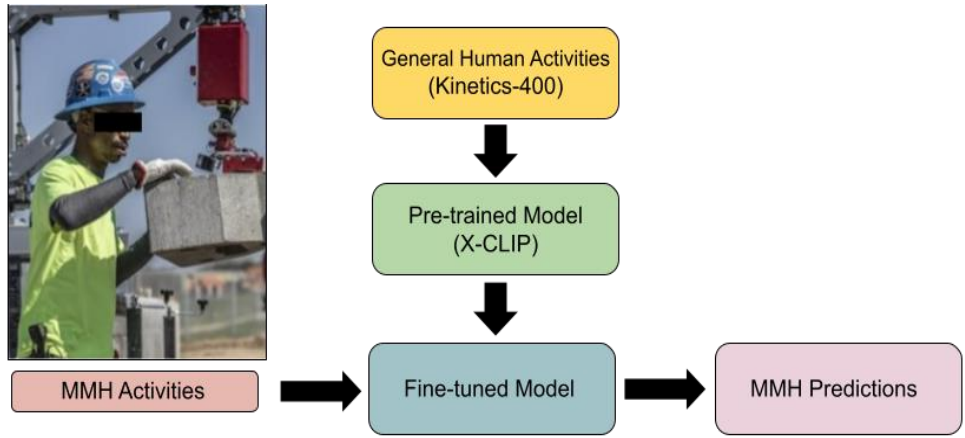

**Figure 1. An overview of the developed methodology (Image courtesy of Construction Robotics (Construction Robotics 2022) and used with permission).**

We sourced the videos from YouTube, selecting 65 videos covering four distinct MMH tasks. The videos include scenarios with and without collaborative robots. Our preliminary analysis indicated that videos with and without robot collaboration do not have a significant impact on the classification accuracy.

To label the videos, we manually annotated each video with the corresponding activity. We defined a set of four MMH activities that are common in construction sites: carrying a load, loading a load, pushing a load, and pulling a load. The X-CLIP model we adopted and advanced further requires 32 frames from the video as input, so we randomly sampled 32 frames from a defined start and end point of the activity in the video. The inclusion of start and end points of activities in videos also allowed us to use the same video for multiple examples for our model, increasing the size of the dataset.

The videos had various resolutions and aspect ratios, but on average were 480x360 pixels and had a frame rate of 30 frames per second. We selected the specific portion of the videos where the activity was happening and processed the video into an array of video frames, sampling 32 frames from each video. Because of this sampling technique, we ensure the portion of video selected is no more than 16 seconds long so that the resulting selected frames are no more than half a second apart in real-time. This ensures that the activity is accurately portrayed and that the frames can be interpreted by the model as sequential.

**Preprocessing.** The model processes images only of size 224×224, so before being fed into the model, the images are cropped to appropriate sizes. During training, the model also augments each frame, resizing images, flipping them, and jittering the color values. This is done to help the model generalize better during training. During validation or inference, the model only crops the video to 224x224 and normalizes color values.

**Training the X-CLIP model.** X-CLIP is a state-of-the-art multimodal model that was developed by Microsoft and is trained on a large corpus of text and video pairs from the internet (Ni et al., 2022). It combines the power of natural language processing (NLP) and computer vision to learn joint representations of text and videos. During training, X-CLIP learns to associate a text description with a video, and vice versa, by optimizing a contrastive loss function that maximizes the similarity between positive pairs of text and videos and minimizes the similarity between negative pairs. Figure 2 shows a schematic overview of the model architecture.

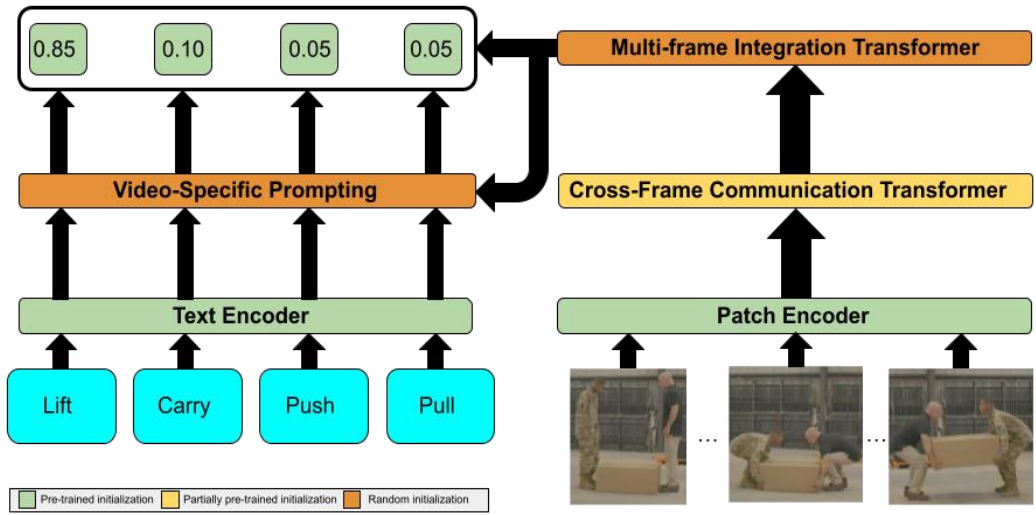

**Figure 2. X-CLIP architecture (Ni et al., 2022).**

The X-CLIP model operates on the principle of contrasting label embeddings and video embeddings using a similarity score. Embeddings are an n-dimensional vector that contains the semantic information necessary to sufficiently capture an input. To attain these embeddings, images and text are passed through their respective encoders and the output is the associated embedding vector. A similarity score indicates the likelihood of a label corresponding to a particular video, with higher similarity scores indicating a higher probability of correspondence. The model employs a combination of encoders and transformers to embed both video and text inputs (Ni et al., 2022). In essence, the model compares video frame embeddings to a set of label embeddings, and uses a transformer to aggregate predictions for each video frame, ultimately outputting the label that most likely corresponds to the video. After collecting and preprocessing the data, we used the X-CLIP model for fine-tuning the activity recognition model on videos of MMH tasks.

**Fine-tuning the X-CLIP model**. Fine-tuning helps the model learn the specific features of the construction activities and improve its accuracy in recognizing these activities. To adapt X-CLIP for MMH as opposed to generic activity recognition, we fine-tuned the model on our dataset of construction MMH videos. Fine-tuning involves updating the weights of the pre-trained model marginally, using our MMH training data to learn better representations that are specific to our task. The innovation of this methodology lies in the fact that for any downstream HAR task, a small team can leverage the power of the pretrained model for any specific set of activities.

We utilized the pre-trained model to reduce training time and computational costs as well as to leverage the generic activity recognition knowledge the model learns in pre-training. The pre-trained model generalizes to classes it did not see during training due to the model learning a label-agnostic embedding space. This allows us to create our own dataset of videos with labels specific to our use case. For this paper, we analyze common MMH tasks. The model is fine-tuned by using a very low learning rate of $\alpha = 10^{-6}$ and training the model on very few examples (~8 videos per class) for 5 epochs with a batch size of 8. Because the model is pre-trained, it extracts relevant visual and semantic information from the beginning of the fine-tuning process and only requires small updates to the weights to learn a better representation of both the label and accompanying videos without the need for many examples of new activities.

**Evaluation**. The model's performance was evaluated using a held-out split of our dataset of construction MMH activities, consisting of 60% of the total data. We report accuracy, precision, and recall metrics for each experiment, and train using cross-entropy loss (Haurilet et al., 2019). The dataset includes activities that the model was not pre-trained on to assess the model's ability to recognize new activities after fine-tuning. The pre-trained model was trained on Kinetics-400, a popular human activity recognition benchmark dataset with 400 distinct activities. Because of the model's use of label embeddings rather than strict activity labels, the MMH activities we selected are not exactly present in the Kinetics-400 dataset, but more general labels are present in the dataset that capture the general idea of the activities we tasked the model to classify. An example of this is that in Kinetics-400, there are labels such as deadlifting or lifting a hat, while we used the general label of lifting materials.

**RESULTS AND DISCUSSION**

The model was fine-tuned to recognize four distinct MMH tasks: carrying a load, loading a load, pushing a load, and pulling a load. We labeled each video with the activity that was being performed and compared the predicted activity to the ground truth.

The overall accuracy of the model, (i.e., the number of correct predictions divided by total predictions on the test set) in recognizing the four activities improved from 46% to 69% (Table 1) after fine-tuning the model for 5 epochs with a learning rate of $\alpha = 10^{-6}$.

**Table 1. Model metrics before and after fine-tuning.**

|  | Precision | Recall | F1 Score | Accuracy |
|---|---|---|---|---|
| Pre-trained | 0.37 | 0.46 | 0.41 | 46% |
| Fine-tuned | 0.74 | 0.69 | 0.72 | 69% |
| Fine-tuned (pushing and pulling combined) | 0.91 | 0.90 | 0.90 | 90% |

In Table 1, we show key model metrics to evaluate the performance of the model before and after fine-tuning. Since the activities of pushing and pulling are mistaken for one another, we also tested the model when combining the activities under one label, "pushing or pulling," and observed an increase in accuracy from 69% to 90%. These results show three things: first that X-CLIP has a robust understanding of human activities and can achieve 46% accuracy on MMH activities it was not pre-trained on, second that fine-tuning the model for a desired activity increases classification accuracy with very few training examples, and third that though it was pre-trained on 400 activities, there are blind spots and activities the model will perform poorly on after fine-tuning (e.g. pulling).

These results demonstrate the effectiveness of our TL methodology using X-CLIP fine-tuning for activity recognition in the construction industry. The ability to recognize MMH activities in real-time can enable the deployment of collaborative robots to assist workers in such activities, leading to increased efficiency and safety in construction sites.

X-CLIP was pretrained on Kinetics-400, and of the 400 activities, there are a few labels similar to our own. Namely, deadlifting, lifting a hat, carrying weight, pushing a cart, and pushing a wheelbarrow. While these labels are not directly related to the MMH activities we analyze in this paper, it can be seen why the model performs better at classifying activities similar to these, since that is what the original model was trained to do on a very large dataset. The only labels related to pulling in Kinetics-400 are "pull ups" and "pulling espresso shot," both of which use the word pull in a different context than we use it in for MMH tasks. Because it was not trained to understand this activity, fine-tuning is not enough to increase the accuracy of pulling classification.

An important point to make here is that the textual content of the label is significant in the video labels and can affect model accuracy by up to 21% (Table 2). X-CLIP works by learning an embedding vector that contains the meaning of the label and learning a different embedding vector for the video content. These vectors are compared by means of a cosine similarity score, being trained to maximize the similarity between the label embedding corresponding to the activity in the video. This means that different labels give different accuracies. In Table 2, we show a few different combinations of labels and the resulting accuracies obtained after fine-tuning the model on the given labels and corresponding videos.

**Table 2. Model accuracy as a function of labels.**

|  | Label 1 | Label 2 | Label 3 | Accuracy |
|---|---|---|---|---|
| Baseline | Lifting a load | Carrying a load | Pushing or pulling a load | 72% |
| Variation 1 | Lifting a box | Carrying a box | Pushing or pulling a box | 82% |
| Variation 2 | Lifting | Carrying | Pushing or pulling | 69% |
| Variation 3 | A photo of someone lifting a box up | A photo of someone carrying materials | A photo of someone pushing or pulling a box | **90%** |

For the baseline, we see much lower accuracy which is related to the model not being familiar with the meaning of the word "load" as it is paired with the accompanying videos. X-CLIP utilizes CLIP's original text encoder which was trained on a vast corpus of image-text pairs of which we do not have access to, but we can assume from the millions of pairs, the text encoder built up a sufficient representation of the English language, but there may be gaps such as using "load" as a noun, a more niche meaning of the word. Because of this, it is a poor choice of words for our model's labels. Using the word "box" in place of load is a more likely combination of words to be found in the original CLIP dataset, and the model performs better, despite the word "box" not appearing in the Kinetics-400 dataset. The model extracts roughly the same visual features independently of the labels provided, it is only that the label embeddings are contingent upon the quality of data in CLIP's original image-text pair dataset.

The accuracy of 90% achieved in our study for activity recognition of construction workers interacting with collaborative robots is comparable to some previous studies in the field of construction activity recognition using video data. For instance, a review paper by Sherafat et al. reported accuracies ranging from 54% to 96% for different construction activities using video-based methods (Sherafat et al., 2020). Nevertheless, previous studies did not employ TL and as such, required significantly more data and computational resources for training and used less generalizable model architectures. Fine-tuning large, generalizable models enables teams with small budgets to leverage previous investments in large models on a bespoke downstream task. It is also important to note that previous studies primarily focused on recognizing routine daily human activities, which are often more distinguishable from each other than MMH tasks in construction. In contrast, our study targets the recognition of MMH tasks, which are highly similar and require a more nuanced approach to achieve accurate recognition. Despite this added complexity, our TL methodology achieved a promising result of 90%, highlighting its potential for real-world applications in the construction industry.

## CONCLUSION

In this study, we presented a TL methodology for activity recognition of construction workers interacting with collaborative robots using the X-CLIP model. The proposed methodology fine-tunes a pre-trained model to accurately classify activities in a new environment and was tested using videos captured from construction MMH activities involving workers and robots. The results showed that the system could successfully recognize distinct MMH tasks, even though such activities were absent in the dataset the model was pre-trained on. This indicates the potential of the proposed methodology in recognizing a wide range of construction activities with high accuracy, which is imperative for the widespread deployment of collaborative robots in construction. The proposed TL methodology utilizing the X-CLIP model provides a promising approach for activity recognition in the construction industry. However, there are limitations that our team is working to address in future work. Further research is needed to evaluate the effectiveness of the proposed methodology in different construction environments and to assess its ability to adapt to new activity types and subjects. Future research can also explore ways of improving the accuracy and robustness of activity recognition models in construction, including using more diverse and larger datasets, more sophisticated feature extraction techniques, and more advanced machine learning algorithms. Finally, different data modalities can help in distinguishing between activities that share similarities in terms of worker body movement.

## ACKNOWLEDGMENT

The presented work has been supported by the U.S. National Science Foundation (NSF) CAREER Award through the grant # CMMI 2047138. The authors gratefully acknowledge the support from the NSF. Any opinions, findings, conclusions, and recommendations expressed in this paper are those of the authors and do not necessarily represent those of the NSF.